# Data-driven Neural Architecture Learning for Financial Time-series Forecasting

Dat Thanh Tran[1], Juho Kanniainen[1], Moncef Gabbouj[1] and Alexandros Iosifidis[2]
[1]Laboratory of Signal Processing, Tampere University of Technology, Finland
[2]Department of Engineering, ECE, Aarhus University, Denmark
{dat.tranthanh, juho.kanniainen, moncef.gabbouj}@tut.fi, alexandros.iosifidis@eng.au.dk

## ABSTRACT

Forecasting based on financial time-series is a challenging task since most real-world data exhibits nonstationary property and nonlinear dependencies. In addition, different data modalities often embed different nonlinear relationships which are difficult to capture by human-designed models. To tackle the supervised learning task in financial time-series prediction, we propose the application of a recently formulated algorithm that adaptively learns a mapping function, realized by a heterogeneous neural architecture composing of Generalized Operational Perceptron, given a set of labeled data. With a modified objective function, the proposed algorithm can accommodate the frequently observed imbalanced data distribution problem. Experiments on a large-scale Limit Order Book dataset demonstrate that the proposed algorithm outperforms related algorithms, including tensor-based methods which have access to a broader set of input information.

## 1. INTRODUCTION

The erratic, dynamic nature of the market and the availability of massive amount of data provide the analysts both opportunities and challenges. The problem is even more challenging in High-Frequency Trading (HFT), which typically involves rapid and complex movements of data. Several mathematical models have been proposed in the past to simulate certain properties of the financial market and to predict asset price, stock trends and so on. The traditional mathematical models make many assumptions of the underlying process that generates the data, which are usually unrealistic in practical cases. With the advances in computing hardware and a massive amount of data aggregated, more and more complex models that impose fewer assumptions and leverage the modern computational power have been proposed recently, e.g., [11, 13, 14].

Nowadays, practitioners are moving from traditional autoregressive models such as Autoregressive Integrated Moving Average (ARIMA) [10] towards ensembles of regression trees [18] or artificial neural networks [11, 13, 14]. In fact, statistical machines have been shown to outperform ARIMA model in different scenarios [5]. While neural network-based solutions can cover a richer set of transformations compared to traditional models and allow low-cost inference, the network designs are often based on heuristics, thus virtually imposing fixed functional forms on different problems. On the other hand, ensemble methods such as Random Forest [7] possess no such limitation by aggregating a collection of weak classifiers that are discovered based on the problem at hand. However, an ensemble of classifiers requires huge operating cost during inference.

In financial forecasting, different data sources coming from different markets or stocks often possess different nonlinear relationships, thus, requiring different transformations. In fact, this is true for many application areas. To tackle the aforementioned problem while taking advantage of neural network-based solution, several works have been proposed to automatically learn the network topology in other application domains [1, 3, 12, 19, 20].

Following similar motivation, in this work, we adapt the recently proposed Heterogeneous Multilayer Generalized Operational Perceptron (HeMLGOP) algorithm [19] to progressively learn a heterogeneous neural architecture for the given financial forecasting problem with potential target imbalance problem. The adaptation modifies the Mean Squared Error (MSE) contributed by different target classes to prevent HeMLGOP from learning network architectures that are biased towards majority classes. In fact, tackling the imbalanced data distribution problem has previously been shown to improve the performance of the financial forecasting system [17]. As indicated by the name, HeMLGOP utilizes Generalized Operational Perceptron (GOP) as the neuron model, which was designed to encapsulate a wide range of nonlinear transformations and shown to surpass the learning capacity of traditional McCulloch-Pitts model [6].

The remaining of our paper is organized as follows: in Section 2, we will review the Generalized Operational Perceptron model and other related progressive neural architecture learning algorithms. In Section 3, we will present the modified HeMLGOP algorithm, followed by the experiments in Section 4. We conclude our work in Section 5.

## 2. RELATED WORKS

Generalized Operational Perceptron (GOP) is a neuron model that was proposed in [6]. The main idea of GOP is to achieve a better simulation of biological neurons observed in mammals by expressing the transformation induced by a neuron in three distinct operations: nodal, pooling and activation operation. Let $x_k$ be the inputs ($k = 1, \dots, K$), and $\psi$, $\rho$ and $f$ be the nodal, pooling and activation operator of a GOP, which sequentially performs the following operations:

$$y_k = \psi(x_k, w_k) \qquad (1)$$

$$z = \rho(y_1, \dots, y_K) + b \qquad (2)$$

$$t = f(z) \qquad (3)$$

where $w_k$ and $b$ denote the adjustable synaptic weight and bias term. In short, nodal operation modifies the incoming signals by using the synaptic weights. Pooling operation summarizes the modified signal, incorporating also the bias term and activation operation performs a thresholding step.

Each GOP selects its nodal, pooling and activation operator from a pre-defined set of operators, i.e., $\psi \in \Psi, \rho \in \mathrm{P}, f \in F$. An example of the set of operators can be found in [6]. In this paper, the term *operator set* refers to a particular combination of nodal, pooling and activation operator. By learning the operator set assignment and its weights based on the given data, an algorithm using GOPs can generate a problem-specific architecture. The authors of GOP proposed an algorithm called Progressive Operational Perceptron (POP), which is computationally intensive for large-scale datasets. For interested readers, details of POP are given in [6].

Similar attempts have been made to learn fully-connected, feedforward networks based on the traditional perceptron or radial basis function such as Stacked Extreme Learning Machine (S-ELM) [15], Broad Learning System (BLS) [3] or more recently Progressive Learning System (PLS) [1]. The similarity between our algorithm and the above-mentioned ones is the utilization of a well-known randomization process [2]. However, different from S-ELM, BLS or PLN, HeMLGOP algorithm takes advantage of GOPs to build the neural architecture from a richer set of functionals.

## 3. HETEROGENEOUS MULTILAYER GENERALIZED OPERATIONAL PERCEPTRON

To define a problem-specific architecture, HeMLGOP adopts the progressive learning paradigm that gradually extends the network topology by adding a new block of GOPs at each step. The algorithm searches for the suitable operator set assignment at each step, allowing heterogeneity, i.e., a hidden layer can have GOPs with different operator sets.

Given a pre-defined block size, HeMLGOP sequentially adds a new block in the following manner: if the progression in the last hidden layer was not terminated in the previous step, the new block is added to the last hidden layer, taking outputs from the second last hidden layer as inputs. Otherwise, the new block forms a new hidden layer, taking outputs from the last hidden layer as inputs.

The increment in a hidden layer stops when the performance of the network saturates. This is quantified by comparing the rate of performance improvement with a given threshold $\epsilon$. Particularly, the progression in the current hidden layer terminates if

$$\frac{\mathcal{L}_t - \mathcal{L}_{t-1}}{\mathcal{L}_t} < \epsilon \quad (4)$$

where $\mathcal{L}_t, \mathcal{L}_{t-1}$ are the loss value at the current step and previous step, respectively. The criterion in (4) is checked after the new block is fully learned, i.e., the suitable operator set has been selected and its weights optimized.

When the current hidden layer is fully grown, its inclusion in the final topology is evaluated. The idea is that, after some steps, the performance of the network gets saturated and we want to stop the progressive learning procedure and proceed to fine-tune all the weights in the network through backpropagation while keeping all the operator set assignments fixed. HeMLGOP terminates the progressive learning routine when

$$\frac{\mathcal{L}_l - \mathcal{L}_{l-1}}{\mathcal{L}_l} < \epsilon \quad (5)$$

where $\mathcal{L}_l$ and $\mathcal{L}_{l-1}$ denote the loss value obtained with and without the current hidden layer. The given threshold $\epsilon$ in (4) and (5) can be different values to adjust the favor of network depth or width. That is, the lower $\epsilon$ is, the wider or deeper the learned network architecture. It should be noted that the criterion in (5) is only evaluated after (4) is satisfied, i.e., after the current hidden layer is fully learned.

The optimization of a new block involves two main steps: the search for the suitable operator set assignment and the weight update through BP after fixing the operator set. Once a new block is optimized, its operator set and weights are fixed. Thus, during the optimization of the new block, all previous blocks' weights and operator sets are fixed. HeMLGOP constrains all GOPs within a block to share the same operator set assignment. In addition, the output layer is a linear layer which takes outputs from the last hidden layer as inputs. The entire output layer is re-calculated at each step, in conjunction with the new block.

To select the best operator set, it is necessary that all combinations of nodal, pooling and activation operators are evaluated. HeMLGOP performs the evaluation by a randomized approach: for each operator set assigned to the new block, random weights drawn from a uniform distribution are assigned to the new block and the output layer weights are obtained by optimizing the re-weighted least-squared problem. To balance the contribution of different target classes, the mean squared error term of a training sample is scaled by a coefficient $s_i$, which is inversely proportional to the popularity of the class the sample belongs to.

Specifically, denote $H \in \mathbb{R}^{N \times D}$ and $Y \in \mathbb{R}^{N \times C}$ the last hidden layer output and the target outputs, the output layer weights $W$ are calculated according to the following formula:

$$W = (H^T S^T S H + \lambda I)^{-1} H^T S^T S Y \quad (6)$$

where $\lambda$ is a hyperparameter that controls the amount of regularization, $I$ is the identity matrix. $S$ is a diagonal matrix of size $N$ with the $i$-th diagonal element being $\sqrt{s_i}$. It should be noted that when $s_i$ are equal for all classes, (6) becomes the standard least-squared solution.

After each operator set is evaluated as described above, the one with the highest performance is assigned to the new block. The new block weights and the output layer weights are further updated for some backpropagation epochs.

While HeMLGOP assumes that the functional form of the new block, i.e., the operator set, can be found via the randomized process, the weights fine-tuning step is necessary to fully harness GOPs in the new block, thus avoiding the redundancy of "weak" neurons. Moreover, after network progression terminates, all weights and biases are further fine-tuned via backpropagation.

## 4. EXPERIMENTS

Experiments were conducted on a large-scale Limit Order Book (LOB) dataset to evaluate the proposed algorithm and other related ones. The next subsection describes the problem of predicting the mid-price movement, followed by the description of hyperparameter settings, and finally experiment results and discussion.

### 4.1 FI-2010 dataset

Limit order is a type of order to buy or sell a certain amount of security at a specified price. In a limit order, the type (buy/sell), the price and the respective volume must be specified. Buy (bid) and sell (ask) limit orders constitute two sides of the Limit Order Book (LOB). At each time instance, mid-price is defined as the mean between the best bid price and best ask price. This quantity is a virtual price, which lies between the best bid and best ask price and its movement reflects the dynamic of the LOB and the market.

Thus, the ability to predict the movement of mid-price in the future plays an important role in analyzing the market. For more information about LOB, we refer [4]. Based on the current best bid and ask orders, we evaluate the performance of all algorithms on the task of predicting the movement of mid-price in the future.

FI-2010 contains more than 4 million limit orders coming from 5 different Finnish stocks during the period of 10 working days provided by Nasdaq Nordic [8]. The database provides 144-dimensional feature vector summarizing information in every block of 10 order events. Each feature vector is associated with the mid-price movements (decreasing, stationary, increasing) in the next $H = \{10, 20, 30, 50, 100\}$ order events. In our experiments, $H = 10$ was used. The database also includes 9-fold anchored forward cross-validation splits on a day basis. In the K-th fold, the first K days are used as training data and the next day is used as the test data.

Since the dataset is imbalanced, the average F1 score is used as the main metric. In addition to F1 score, we also report accuracy, average precision and recall score.

## 4.2 Hyperparameters

We conducted experiments with HeMLGOP, S-ELM [15], BLS [3] and PLN [1]. In addition, we also include previous results reported for the following algorithms: Ridge Regression (RR), Single Layer Feedforward Network (SLFN) [8], Linear Discriminant Analysis (LDA), Multilinear Discriminant Analysis (MDA), Multilinear Class-specific Discriminant Analysis (MCSDA) [16], Multilinear Tensor Regression (MTR), Weighted Multilinear Tensor Regression (WMTR) [17], Bag-of-Feature (BoF) and Neural Bag-of-Feature (N-BoF) [9].

We should note that HeMLGOP, S-ELM, BLS, PLN, RR, SLFN, and LDA operates on vector inputs, taking only the 10 most recent order events information. On the contrary, other methods operate on tensor inputs, utilizing extensive past information (at least 100 order events to make predictions. Methods which take advantage of tensor representation are abbreviated with an asterisk in the result table.

For S-ELM, BLS and PLN, all the regularization parameters were selected from the set $\{10^{-3}, 10^{-2}, 10^{-1}, 1, 10, 10^2, 10^3\}$. The number of iterations in ADMM in BLS and PLN was set to 500. For BLS and PLN, each layer starts with 100 neurons and increments by a step of 50 to maximum 1000 neurons. For S-ELM, the hidden layer and PCA dimension were set to 1000 and 500 respectively.

Regarding HeMLGOP, the same set of operators in [19] was used in our experiments. The block size was set to 40 and the maximum number of blocks per layer and the maximum number of layers were 4 and 8 respectively. During the progression, new block weights were updated for 300 backpropagation epochs, with the initial learning rate of 0.01, which was reduced by 0.1 every 100 epochs. $\lambda = 1.0$ and either weight decay (0.0001) or max-norm (3.0) was used to regularize our networks during backpropagation. A similar setting was applied during the final network fine-tuning. Since FI-2010 is an imbalanced dataset, the weight of each class in Eq. (6) was set to be reversely proportional to the number of samples of that class in the training set.

The same termination criterion as in (4) and (5) were applied for all progressive learning algorithms. For other algorithms, experiment settings can be found from the original works.

## 4.3 Result

Table 1 shows the average performance of all competing algorithms over 9 folds. It is clear that HeMLGOP performs best in terms of F1 score, which reflects the trade-off between precision and recall. Other progressive learning algorithms (S-ELM, BLS, and PLN) achieve higher accuracy scores, however with very low recall rates. Since the data distribution among different movements is highly imbalanced, accuracy does not reflect the quality of a model. By re-weighing the loss contributed by different classes, HeMLGOP outperforms all other vector-based methods by a relatively large margin, nearly 8% difference in average F1 compared to the second-best algorithm (N-BoF).

Compared with tensor-based approaches that are designed to take advantage of a long sequence of past information, HeMLGOP still establishes a gap of almost 3% difference in F1 score. It is worth noting that MDA, MCSDA and WMTR were also formulated to take into account the class imbalance problem.

**Table 1. Prediction performance (%) on FI-2010. Asterisk (*) denotes methods operating on tensor inputs**

| Models | Accuracy | Precision | Recall | F1 |
|---|---|---|---|---|
| Prediction Horizon H=10 | | | | |
| RR [8] | 48.00 | 41.80 | 43.50 | 41.00 |
| SLFN [8] | 64.30 | 51.20 | 36.60 | 32.70 |
| LDA [16] | 63.82 | 37.93 | 45.80 | 36.28 |
| S-ELM [15] | 89.34 | 55.17 | 34.53 | 33.94 |
| BLS [3] | 89.46 | 63.22 | 33.68 | 32.20 |
| PLN [1] | 87.65 | 47.60 | 39.36 | 40.64 |
| BoF [9] | 57.59 | 39.26 | 51.44 | 36.28 |
| N-BoF [9] | 62.70 | 42.28 | 61.41 | 41.63 |
| MDA* [16] | 71.92 | 44.21 | 60.07 | 46.06 |
| MCSDA* [16] | 83.66 | 46.11 | 48.00 | 46.72 |
| MTR* [17] | 86.08 | 51.68 | 40.81 | 40.14 |
| WMTR* [17] | 81.89 | 46.25 | 51.29 | 47.87 |
| HeMLGOP | 83.06 | 48.57 | 50.67 | **49.43** |

## 5. CONCLUSION

In this work, we proposed a method for mid-price movement prediction based on limit order book data. The proposed approach is based on an algorithm that automatically determines the neural network architecture for financial time-series prediction based on training data. Empirical result shows that the proposed method outperforms related ones, including multilinear methods that utilize privileged information.